\documentclass[11pt]{article}
\usepackage[utf8]{inputenc}
\usepackage[a4paper,margin=2.5cm]{geometry}
\usepackage{amsmath,amssymb}
\usepackage{graphicx}
\usepackage{booktabs}
\usepackage{multirow}
\usepackage{xcolor}
\usepackage{enumitem}
\usepackage[font=small,labelfont=bf]{caption}
\usepackage[expansion=false]{microtype}
\usepackage{tikz}
\usetikzlibrary{arrows.meta,positioning,fit,calc,shapes.geometric}
\usepackage[round,authoryear]{natbib}
\usepackage[colorlinks=true,linkcolor=blue!50!black,citecolor=blue!50!black,urlcolor=blue!50!black]{hyperref}
\usepackage[title]{appendix}

\definecolor{ink2}{HTML}{52514E}
\definecolor{blu}{HTML}{2A78D6}
\definecolor{org}{HTML}{EB6834}
\definecolor{grn}{HTML}{1BAF7A}

\setlist{itemsep=2pt,topsep=4pt}

\title{Knowing is not using:\\ Controlling reactivity in a repeated trust game}
\author{Ver\'onica Sanz\\[2pt]
  \small Instituto de F\'isica Corpuscular (IFIC), Universitat de Val\`encia--CSIC}
\date{\small Draft, 30 November 2026}
\begin{document}
\maketitle

\begin{abstract}
An internal representation can track a changing environment accurately without driving a commensurate change in behavior. Here, we demonstrate this decoupling in a repeated trust game and show how targeted interventions increase behavioral reactivity by unlocking the policy's use of that latent information. Specifically, we train a small language model via reinforcement learning to play a 20-round repeated trust game with a hidden, switching partner type. While reward training aligns linear readouts of internal activations with the exact Bayesian posterior, the policy's behavioral reactivity depends strongly on the volatility of the training environment. Agents trained on stationary partners exhibit strong hysteresis (lock-in) following a mid-game partner switch, even though their residual stream tracks the new partner type nearly as accurately as agents trained on switching partners ($0.84$ vs.\ $0.86$ normalized decoder response). We map the policy's choice dynamics onto a one-parameter constant-hazard Bayesian filter, in which an effective hazard rate $\hat{h}$ sets a saturation ceiling $\pm \ln[(1-h)/h]$ on accumulated log-odds, bounding the memory timescale of past evidence. Across 25 agents trained at five hazard rates, the inferred $\hat{h}$ increases monotonically with $h_{\mathrm{train}}$ in every seed, though nonlinearly. By varying how often partners switch during training, we tune how readily the learned policy responds to new evidence. Training volatility thus acts as a control parameter that tunes the effective memory horizon of the policy without altering the fidelity of its internal state representation.
\end{abstract}

\section{Introduction}

Suppose an agent has invested repeatedly with a good partner in a non-equilibrium sequential decision process, and the returns suddenly fall. How much new evidence does it take for the agent to stop investing? The answer depends on the effective dynamical model the agent has internalized about environmental volatility. If partners remain stationary, a poor return should count only weakly against a long record of good ones. If partners switch frequently, the same return justifies a rapid revision.

\begin{figure}[t]
\centering
\begin{tikzpicture}[
  font=\small, >={Stealth[length=2mm]},
  box/.style={draw=ink2, rounded corners=2pt, align=center, inner sep=4pt, fill=white},
  lay/.style={draw=ink2, fill=gray!8, minimum width=2.2cm, minimum height=0.4cm, inner sep=1pt, font=\scriptsize},
  probe/.style={rounded corners=2pt, align=center, inner sep=3pt, font=\scriptsize},
]
\node[box, text width=4.1cm, font=\tiny\ttfamily, align=left] (hist) at (1.9,0.35) {Repeated trust game with partner\\ Xafo. Each round choose A (invest)\\ or B (keep).\\[2pt]
Round 1 | action: A | return: 1.62\\ Round 2 | action: A | return: 0.87\\ Round 3 | action: B | kept: 1.00\\ Round 4 | action:};
\node[lay] (l28) at (6.3,1.45) {layer 28};
\node[lay] (l24) at (6.3,0.9) {layer 24};
\node[font=\scriptsize] (d1) at (6.3,0.4) {$\vdots$};
\node[lay] (l14) at (6.3,-0.15) {layer 14};
\node[font=\scriptsize] (d2) at (6.3,-0.65) {$\vdots$};
\node[draw=ink2, dashed, rounded corners=3pt, fit=(l28)(d2), inner sep=4pt, label={[font=\scriptsize]above:language model (Qwen3)}] (lm) {};
\draw[->] (hist.east) -- (lm.west |- hist.east);
\node[box, font=\scriptsize] (act) at (11.0,1.45) {choice: A (invest)\\ or B (keep)};
\draw[->] (l28.east) -- (act.west);
\node[box, font=\scriptsize] (partner) at (11.0,-0.35) {partner of hidden type:\\ returns $3X$, mean of $X$ is $c$};
\draw[->] (act) -- (partner);
\draw[->, ink2] (partner.south) -- (11.0,-2.95) -- node[below, font=\scriptsize]{payoff is added to the history; reward trains the model} (1.9,-2.95) -- (hist.south);
\node[probe, draw=blu, fill=blu!6] (dec) at (4.3,-1.95) {\textbf{(1) read-out}: decoder\\ ``good or bad partner?''};
\draw[->, blu] (l14.west) -| (dec.north);
\node[probe, draw=org, fill=org!6] (steer) at (8.4,-1.6) {\textbf{(2) intervention}: push\\ along the decoded direction};
\draw[->, org] (steer.north) |- (l14.east);
\node[probe, draw=grn, fill=grn!6] (head) at (9.3,2.8) {\textbf{(3) decision head}: narrow\\ channel from one layer to a choice};
\draw[->, grn] (l24.east) -| (head.south);
\draw[->, grn, dashed] (head.east) -| (act.north);
\end{tikzpicture}
\caption{\textbf{The repeated trust game.} The model reads the transcript and chooses whether to invest. Reward trains the policy. A decoder reads the partner from one layer, a probe intervention perturbs a direction, and a separate head learns to choose from frozen activations.}
\label{fig:setup}
\end{figure}
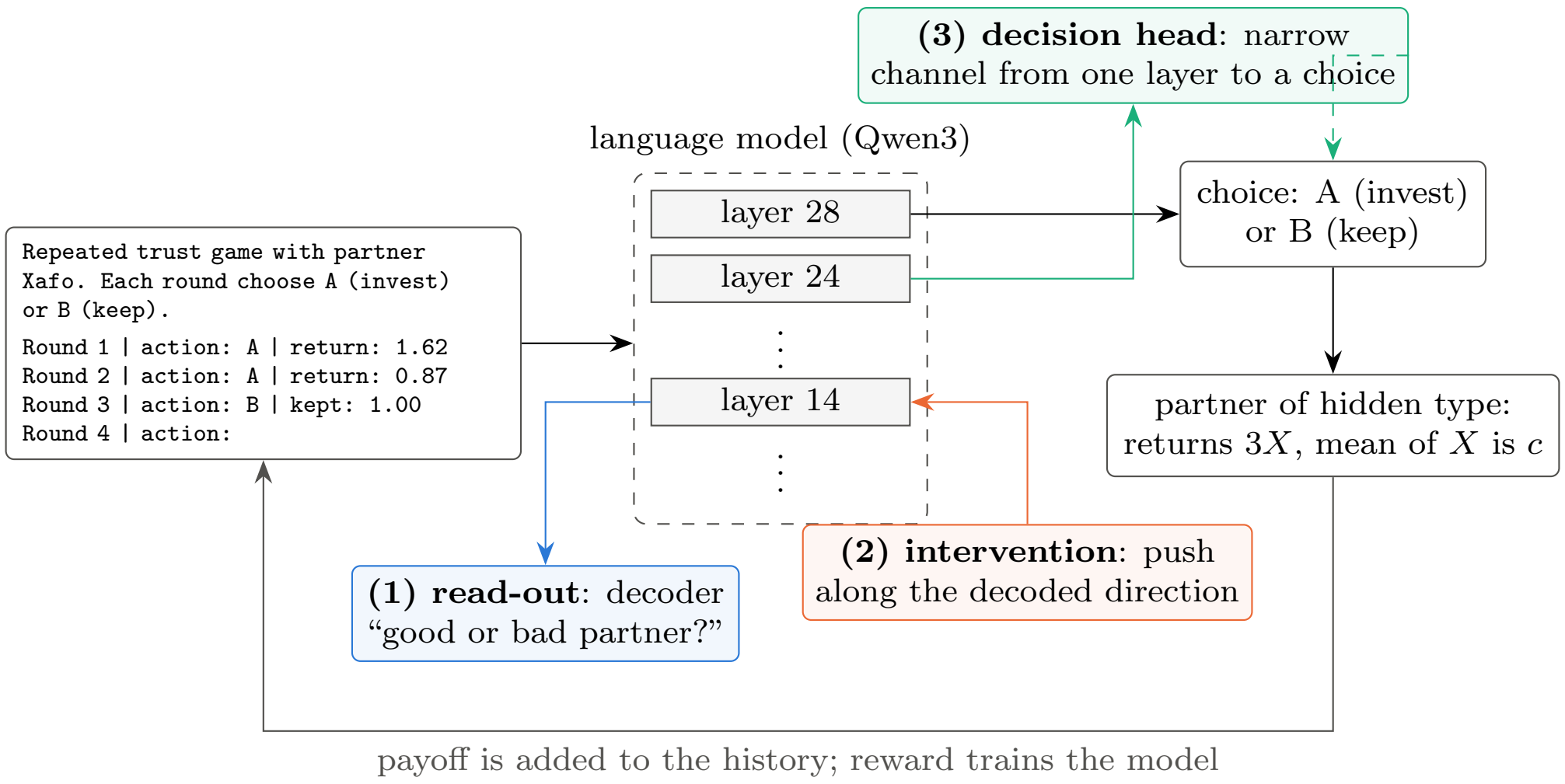

We study this separation of timescales in a repeated trust game \citep{berg1995} played by a small language model. Figure~\ref{fig:setup} shows the closed loop. The task isolates three physical questions: Can reward training teach the network to distinguish hidden partner states? Can fast, high-fidelity state estimation be read out linearly from internal activations even when the policy's behavioral response exhibits slow hysteresis? And can we tune that behavioral memory horizon by varying environmental volatility during training?

Work on probing distinguishes representational decodability from causal use in downstream computation \citep{belinkov2022,ravichander2021,elazar2021,kumar2022}. Work on non-stationary environments relates environmental volatility to the effective learning rate at which new evidence overwrites past beliefs \citep{yu2005,behrens2007,nassar2010}. The repeated game brings these two phenomena into a single, analytically tractable setting with two levels of description. The dependence on training volatility also connects directly to meta-learning via reinforcement \citep{wang2016,duan2016,mikulik2020}. The neural network maps a discrete transcript to a probability of investing. A constant-hazard Bayesian filter provides an effective coarse-grained theory of this map, parameterized by a single effective hazard rate $\hat{h}$ that governs the persistence of accumulated evidence. By varying the true training hazard $h_{\mathrm{train}}$ experimentally and inferring $\hat{h}$ from the resulting choice distributions, we test whether this single parameter captures the policy's response to mid-game switches.

The analytical model was formulated from our initial experiments and subsequently tested on fresh histories and games as well as in an independent sweep over training hazards, with all predictions and decision rules fixed prior to running the evaluation protocols.

\section{A Bayesian model of reactivity}
\label{sec:theory}

In this section we discuss the analytical framework used to describe the experiments. 

We start with the trust game. In each of 20 rounds the agent keeps a coin for a deterministic payoff of $1$ or invests ($a_t=1$) for a stochastic payoff $3X$, where $X$ is drawn from a beta distribution with hidden mean $c$. A partner is good ($G$) if $c>1/3$ and bad ($B$) otherwise. Keeping ($a_t=0$) leaves the partner state unobserved. The held-out test partners have $c_B=1/3-0.15$ and $c_G=1/3+0.15$. Without information about partner type, either fixed action earns $20$ in expectation over 20 rounds, whereas an oracle that knows the instantaneous type earns $24.5$.

Now write $\lambda_t = \log[P(G_t)/P(B_t)]$ for the predictive log-odds of a good partner before round $t$, and $\ell(x)=\log[p(x\mid G)/p(x\mid B)]$ for the log-likelihood ratio supplied by a return $x$. Under a symmetric per-round switch probability (hazard rate) $h \in [0, 1/2)$, exact Bayesian filtering reduces to the one-dimensional nonlinear recursion
\begin{align}
 \lambda_{t+1}&=g_h\!\left(\lambda_t+a_t\ell(x_t)\right), \label{eq:e1}\\
 g_h(u)&=\log\frac{(1-h)e^u+h}{he^u+1-h}. \label{eq:e3}
\end{align}
The inner sum updates the posterior log-odds with the observed return, while the transfer map $g_h(u)$ propagates the belief forward across a potential state transition. The asymptotic behavior of $g_h(u)$ directly controls the memory horizon of the observer:
\begin{equation}
 g_h(u) \approx 
 \begin{cases} 
 \operatorname{sgn}(u)\,\lambda_{\max}(h) & \text{for } |u| \gg \lambda_{\max}(h), \\
 (1 - 2h)\,u & \text{for } |u| \ll 1,
 \end{cases}
 \qquad \lambda_{\max}(h) \equiv \log\frac{1-h}{h}.
 \label{eq:e_asymp}
\end{equation}
The benchmark evaluates Eqs.~\eqref{eq:e1}--\eqref{eq:e3} using the exact return distributions of the two held-out partners.

At $h=0$, $g_0(u) = u$ is the identity map, so $\lambda_t$ undergoes an unbounded, biased random walk (ballistic accumulation). After $m$ investments with a good partner, the expected log-odds grow extensively as $m D_G$, where $D_G=\mathbb E_G[\ell] = D_{\mathrm{KL}}(p_G \parallel p_B) = 7.35\text{ nats}$ is the Kullback--Leibler divergence per observation. Following a switch to a bad partner, the drift per investment reverses to $-D_B=\mathbb E_B[\ell] = -D_{\mathrm{KL}}(p_B \parallel p_G) = -8.80\text{ nats}$. The mean number of contradictory observations required to cross the $\lambda_t = 0$ decision boundary is therefore governed by the ratio of KL divergences:
\begin{equation}
 n^\ast \simeq m \frac{D_G}{D_B} \simeq 0.84\,m. \label{eq:e2}
\end{equation}
The longer the agent has invested with a good partner, the more
contradictory returns it typically needs to reverse its assessment.
This is the origin of lock-in in the stationary filter.

For $0<h<1/2$, $g_h$ acts as a soft-clamping nonlinearity that bounds the predictive log-odds to $|\lambda_{t+1}| \le \lambda_{\max}(h) = \log[(1-h)/h]$. Once accumulated evidence exceeds $\lambda_{\max}(h)$, memory of earlier rounds is erased, replacing extensive accumulation with a finite saturation ceiling. For a myopic decision rule ($a_t = \Theta(\lambda_t)$), a contradictory return reverses an established good assessment immediately when $\ell(x_t) < -\lambda_t$. The single-round reversal probability after a good-to-bad switch is thus a tail probability of $\ell(x_t)$ under $p_B$, evaluated at a threshold bounded by $\lambda_{\max}(h)$. The saturation cap is $6.9\text{ nats}$ at $h=0.001$ and $2.2\text{ nats}$ at $h=0.1$. Because a single typical return from a bad partner supplies $-8.80\text{ nats}$, a single informative draw suffices to overturn even saturated confidence whenever $h>0$. Moreover, during rounds without investment ($a_t=0$), $g_h$ contracts $\lambda_t$ toward zero at rate $(1-2h)$ per step in the linear regime, actively decaying stale beliefs.

Finally, passive state estimation and closed-loop active sampling couple asymmetrically across the $\lambda_t = 0$ boundary. For a deterministic myopic player that invests if and only if $\lambda_t > 0$, the region $\lambda_t \le 0$ acts as an absorbing barrier for information acquisition: once $\lambda_t \le 0$, the agent chooses $a_t = 0$, zeroing out the likelihood channel $a_t \ell(x_t) = 0$. Consequently, $\lambda_t$ relaxes deterministically toward zero from below via $\lambda_{t+1} = g_h(\lambda_t)$ without ever crossing into $\lambda_t > 0$. Recovering after a bad-to-good switch therefore requires stochastic exploration ($P(a_t=1 \mid \lambda_t < 0) > 0$), i.e., a finite effective policy temperature. Thus, good-to-bad switches directly probe the belief-updating timescale while the agent is actively sampling, whereas bad-to-good switches jointly probe belief updating and exploratory re-testing. Table~\ref{tab:dictionary} summarizes how each variable in this filter maps onto the trained language model and our empirical probes.

\begin{table}[t]
\centering\small
\resizebox{\linewidth}{!}{%
\begin{tabular}{lll}
\toprule
Bayesian filter term & RL / network counterpart & Physical / behavioral intuition\\
\midrule
Log-odds belief $\lambda_t$ & Linear belief read-out (layer 24 ridge fit) & Accumulated confidence that the current partner is good vs.\ bad\\
Log-likelihood ratio $\ell(x_t)$ & Observed payoff token ($3X$ in transcript) & Single-round evidence weight; $D_G, D_B$ set the drift rate\\
Action gate $a_t \in \{0,1\}$ & Policy token choice (\texttt{A} = invest, \texttt{B} = keep) & Opens or closes the observation channel ($\lambda_t \le 0$ is absorbing if deterministic)\\
Assumed hazard $h$ & Training switch rate $h_{\mathrm{train}}$ & Volatility prior governing how fast old evidence decays\\
Inferred hazard $\hat{h}$ & Fitted effective hazard from choice logits & Behavioral reactivity coordinate; sets memory ceiling $\lambda_{\max}(\hat{h})$\\
Myopic threshold $\lambda_t > 0$ & Policy softmax / auxiliary decision head & Maps internal belief to action probability; finite gain softens the step\\
\bottomrule
\end{tabular}}
\caption{\textbf{Correspondence between the Bayesian filter and the trained language model.} Separating internal state estimation ($\lambda_t$) from policy readout ($a_t$) and effective memory ($\hat{h}$) isolates where training volatility acts.}
\label{tab:dictionary}
\end{table}

\begin{figure}[t]
\centering
\includegraphics[width=\linewidth]{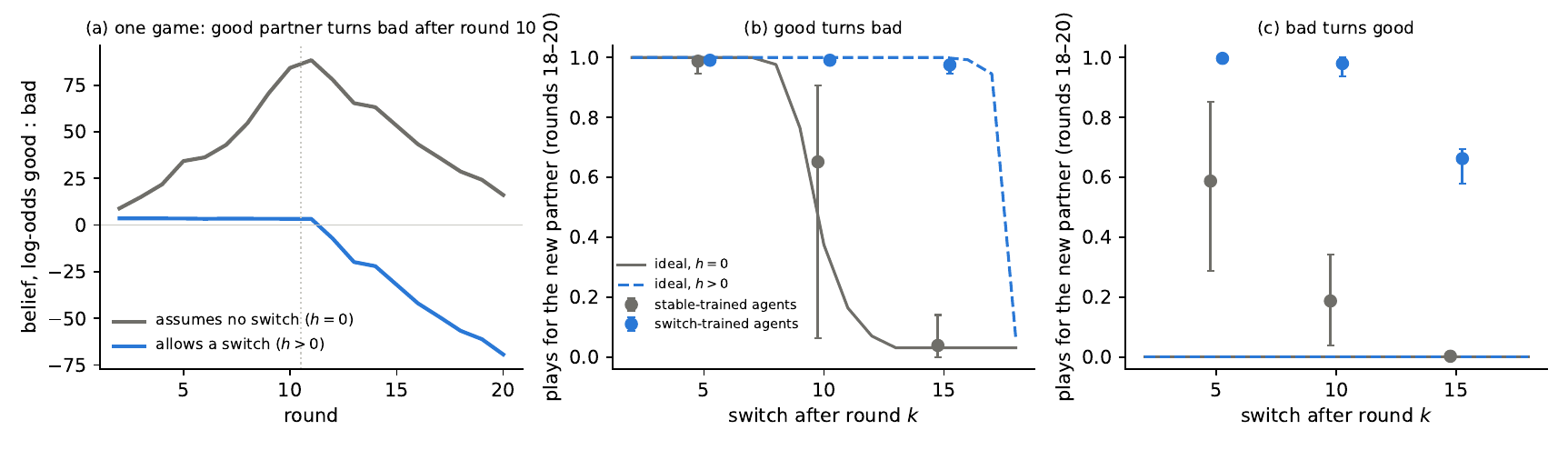}
\caption{\textbf{Belief and behaviour after a switch.} (a) Bayesian log odds on one always-investing transcript with a good-to-bad switch after round 10. The stationary observer accumulates evidence; the one-switch reference can reverse rapidly. (b,c) Fraction of games in which late play favours the new partner, for the two myopic references and the trained agents. In (c), the myopic references stop investing and therefore miss the recovery. Points show means and ranges over five seeds. These data preceded the theory tests.}
\label{fig:theory}
\end{figure}
\section{Learning to play and learning to read}
\label{sec:obs1}

We train Qwen3-0.6B \citep{qwen3} using REINFORCE \citep{williams1992} and a rank-64 low-rank adapter \citep{hu2021lora}. The model processes the discrete game transcript and outputs probabilities over the two action tokens. Training uses four return means symmetrically flanking break-even, and evaluation uses two interpolating means absent from training. Each training condition comprises five independent random seeds. First, we train with stationary partners ($h_{\mathrm{train}}=0$) and compare against the pretrained base model and a control trained on reward-scrambled trajectories.

\begin{figure}[t]
\centering
\includegraphics[width=\linewidth]{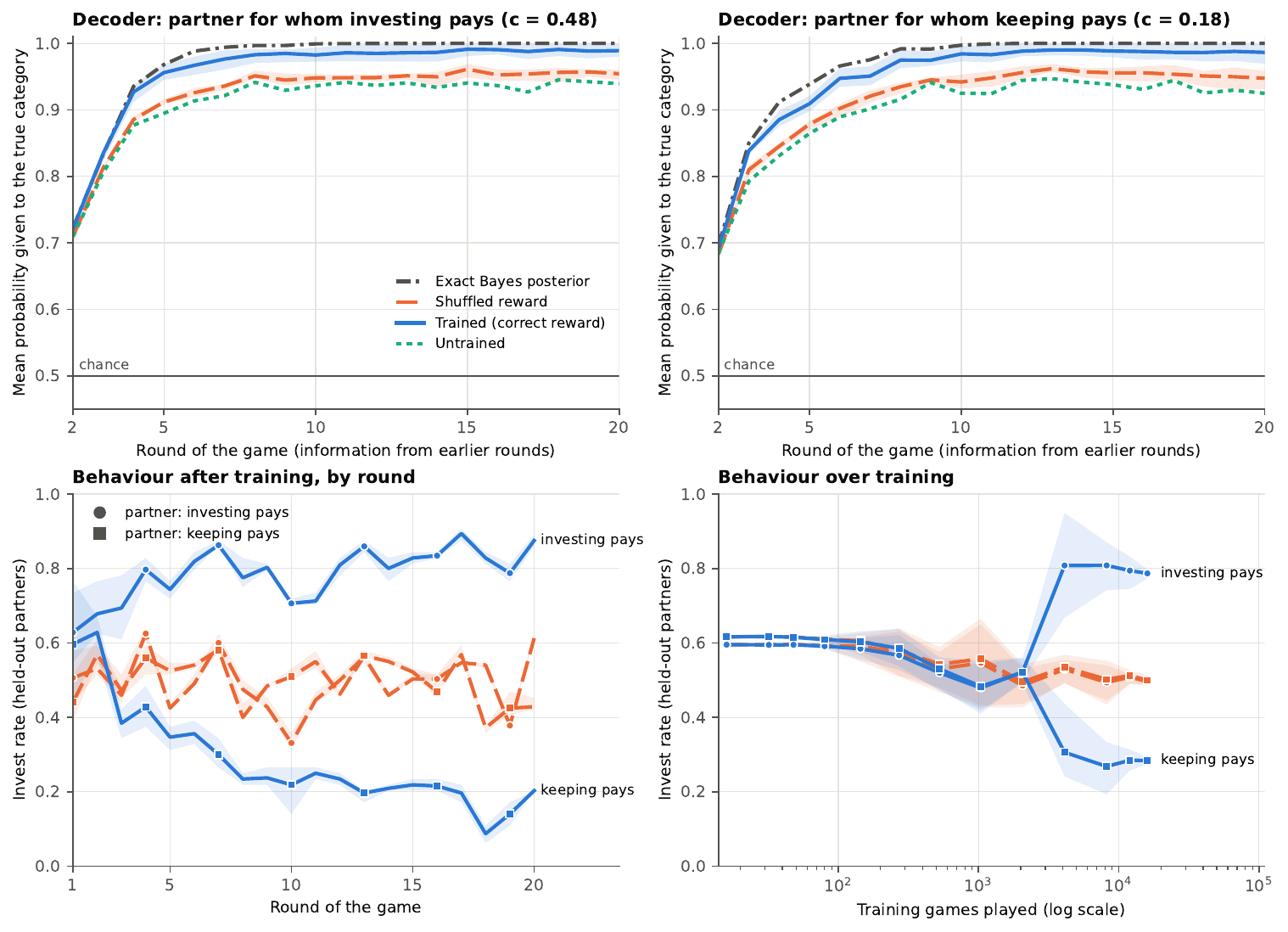}
\caption{\textbf{How the model learns who to trust} (Qwen3-0.6B, five seeds; line: mean, band: range across seeds). \emph{Top:} probability that the layer-14 decoder gives to the true partner type at each round, on the same held-out histories for every model, with the exact Bayesian posterior as the ceiling. \emph{Bottom left:} the trained models' own play against the two test partners. \emph{Bottom right:} play over training; the first point is the untrained model.}
\label{fig:b1}
\end{figure}

The stationary-trained agent invests in 79\% of rounds with a good partner and 28\% with a bad one, earning $22.3$ per game compared with approximately $20$ for both controls. It invests in 61\% of initial rounds and separates the two partner types within a few rounds (Fig.~\ref{fig:b1}). The policy thus learns to sample returns and condition its actions on the inferred partner type.

To measure how accurately partner identity is represented internally independent of the policy's own action-selection bias, we fit a linear \emph{stationary-type probe} to layer-14 residual-stream activations on common random-action transcripts. Held-out log-loss drops to $0.068\text{ nats}$ per decision after reward training, down from $0.124\text{ nats}$ before training and $0.107\text{ nats}$ with scrambled rewards, compared with the exact Bayesian posterior floor of $0.048\text{ nats}$. Reward training thus closes 74\% of the initial gap to the Bayes-optimal bound. At 1.7B parameters, the decoding log-loss improves from $0.084$ before training to $0.065$ after training (against a $0.047$ posterior floor, closing 50\% of the gap). All remaining experiments use the 0.6B model.

\section{Readable evidence, different responses}
\label{sec:obs3}

To decouple internal state estimation from behavioral policy reactivity, we evaluate four distinct observables across two training regimes: \emph{stationary-trained} agents ($h_{\mathrm{train}}=0$) and \emph{switch-trained} agents (trained with a single switch in half of all episodes). For both groups, we measure: (1) the \emph{stationary-type probe} (trained on $h=0$ histories), (2) a \emph{current-type probe} (trained on switching histories to decode the instantaneous partner type $c_t$), (3) a \emph{belief readout} (ridge regression mapping activations onto the ideal Bayesian posterior), and (4) the \emph{behavioral policy} $P(a_t=1 \mid \text{transcript})$.

Evaluated on novel switch rounds and held-out partners, the behavioral policies diverge sharply. Measuring post-switch play relative to each policy's stationary baseline against the old and new partners, the normalized adaptation index traversed after a switch is $0.27$ for stationary-trained agents versus $0.73$ for switch-trained agents—a behavioral contrast of $0.45$ (95\% bootstrap CI $[0.41, 0.49]$).

\begin{figure}[t]
\centering
\includegraphics[width=0.85\linewidth]{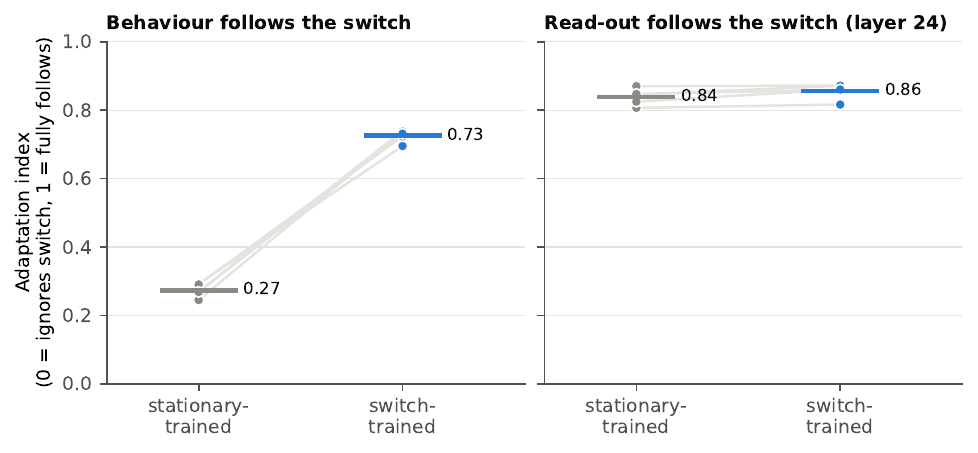}
\caption{\textbf{Policy response and current-type read-out.} Adaptation index after a mid-game partner switch, for stationary-trained (grey) and switch-trained (blue) agents; dots are the five seeds, lines join the same seed, bars are means. \emph{Left:} the agents' own behaviour. \emph{Right:} a linear read-out of the current partner from layer 24 of the same agents.}
\label{fig:knowing}
\end{figure}

In contrast, the current-type probe fitted at layer 24 shows nearly identical representational tracking across the two groups, achieving normalized post-switch responses of $0.84$ for stationary-trained agents and $0.86$ for switch-trained agents (Fig.~\ref{fig:knowing}). Because the probe is evaluated on common random-action transcripts whereas closed-loop play depends on the policy's own sampling choices, we next isolate both effects.

\begin{figure}[h]
\centering
\includegraphics[width=\linewidth]{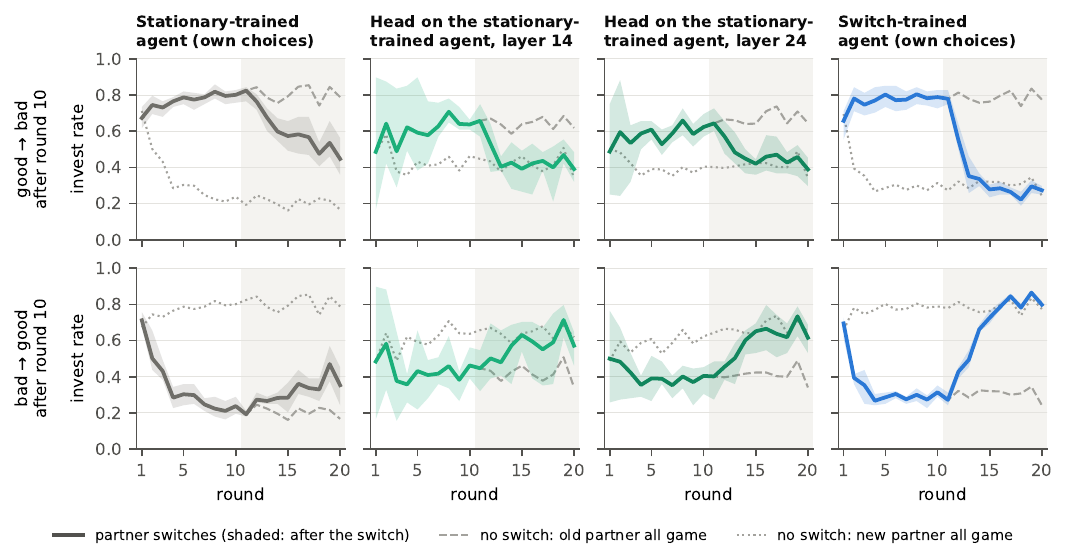}
\caption{\textbf{Behaviour around a partner switch after round 10.} Invest rate per round (mean of five seeds; band: range of the seed means; 128 games per seed). Grey lines are the same agent with no switch, facing the old partner (dashed) or the new one (dotted) all game. A fully adaptive agent would jump from the dashed to the dotted line. The rigid agent stays closer to the old partner; the switch-trained agent moves most of the way. The heads move much of the way to their own new-partner level, but their two reference lines are close together, so the move is small. Descriptive, added during revision, from the stored evaluation games.}
\label{fig:postswitch}
\end{figure}

The behavioral hysteresis persists even when stationary-trained agents do acquire new observations. Conditioning on episodes where a bad partner switches to good after round 10 and the agent invests at least twice post-switch, stationary-trained agents shift their play to favor the new partner in only 33\% of games, whereas switch-trained agents do so in 98\% of games after a single post-switch investment. Moreover, the activations of the stationary-trained network already contain sufficient information to support faster switching: training an auxiliary decision head on frozen layer-14 or layer-24 activations shifts the absolute post-switch investment contrast from $-0.26$ to $+0.08$--$0.09$, compared with $+0.25$ for the full switch-trained policy (Fig.~\ref{fig:postswitch}).

\paragraph{Testing the filter on new histories.}
To remove sampling bias entirely, we evaluate both groups on fresh, common random-action transcripts and fit their action probabilities using either the stationary ($h=0$) Bayesian belief or the single-switch change-point Bayesian belief. On held-out histories, the $h=0$ belief yields a significantly better fit to the behavioral policy of stationary-trained agents in all five seeds, whereas the change-point belief fits the behavioral policy of switch-trained agents better in all five seeds (Table~\ref{tab:bt1}). Remarkably, however, a linear belief readout from layer 24 of the stationary-trained agents reconstructs the \emph{change-point} Bayesian posterior with held-out $R^2 = 0.92$--$0.95$. Thus, the stationary-trained network computes an accurate, volatile state estimate in its residual stream while its policy readout remains governed by an effectively stationary filter.

\begin{table}[h]
\centering\small
\resizebox{\linewidth}{!}{%
\begin{tabular}{llll}
\toprule
& Prediction & Result (five seeds) & Verdict\\
\midrule
T1a & stationary-trained: $h=0$ belief fits better & $\Delta=+0.087$ to $+0.107$; mean $+0.096$ [$+0.091$, $+0.102$] & holds\\
T1b & switch-trained: change-point belief fits better & $\Delta=-0.026$ to $-0.054$; mean $-0.044$ [$-0.052$, $-0.034$] & holds\\
T2 & change-point belief readable (stationary-trained, layer 24) & $R^2=0.92$--$0.95$ & holds\\
T3 & lock-in: stationary-trained notice $k=7$ in $>\frac12$, $k=12$ in $<\frac12$ & 0.89 [0.71, 0.99]; 0.36 [0.12, \textbf{0.60}] & \textbf{fails}\\
& \phantom{lock-in:} switch-trained $\ge0.9$ at both & 0.99, 0.99 & \\
T4 & switch-trained re-test more & $+0.15$ [$+0.08$, $+0.22$] & holds\\
\bottomrule
\end{tabular}}
\caption{\textbf{The test of the theory.} $\Delta$ is the difference in excess cross-entropy between the fits of each agent's probability of investing by the change-point belief and by the $h=0$ belief, on new random-action histories (positive: the $h=0$ belief fits better). Intervals: seed bootstrap (T1, T3) or crossed bootstrap over seeds and games (T4), 10,000 draws.}
\label{tab:bt1}
\end{table}

Four of the five prospectively specified predictions hold, including the elevated exploratory re-testing rate of switch-trained agents ($+0.15$). The quantitative lock-in threshold (T3) fails: for a good-to-bad switch after round 12, an exact $h=0$ Bayesian observer detects the reversal in only $7\%$ of games due to deep ballistic accumulation, whereas stationary-trained agents detect it in $36\%$ of games on average (with two seeds exceeding $50\%$). Physically, this deviation reflects finite-context imperfect integration: a finite-capacity transformer operating at nonzero softmax temperature does not perform lossless summation over 20 rounds, introducing an intrinsic baseline leak rate $h_0 > 0$ that softens the $m D_G$ lock-in barrier.

\section{Tuning reactivity through training}
\label{sec:hz1}

To vary the hazard parameter in Eq.~\eqref{eq:e3} directly at training time, we train 25 fresh agents across $h_{\mathrm{train}} \in \{0, 0.02, 0.05, 0.10, 0.20\}$, pairing five seeds across switch rates. At each round boundary, the training partner switches category with probability $h_{\mathrm{train}}$, allowing multiple transitions within a 20-round episode. For each trained agent, we infer a single effective hazard $\hat{h}$ on fresh random-action transcripts by sweeping over candidate filters and fitting the policy's action log-odds as an affine function of the clipped belief log-odds; $\hat{h}$ is the parameter minimizing the held-out cross-entropy loss. We then evaluate a myopic Bayesian observer equipped with each agent's own $\hat{h}$ on separate switching games.

\begin{figure}[t]
\centering
\includegraphics[width=\linewidth]{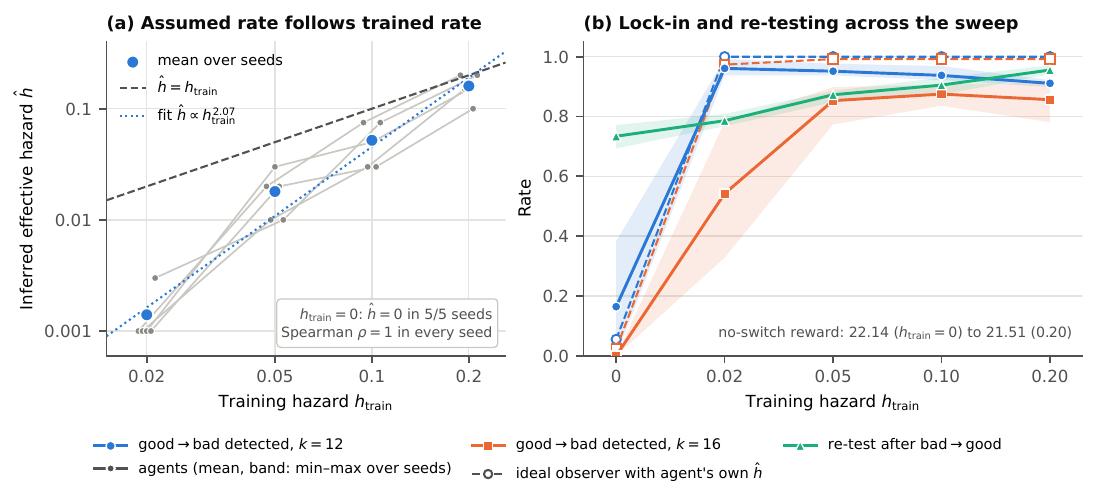}
\caption{\textbf{Tuning reactivity through training volatility} (25 agents across five training hazards $h_{\mathrm{train}}$ and five paired seeds; shaded bands show min--max range across seeds). \emph{(a)} Effective hazard $\hat{h}$ inferred from choice logits on random-action histories versus $h_{\mathrm{train}}$ on log--log axes. Ordering is strictly monotonic in all 5/5 seeds ($\rho=1$, $H1$), while finite-horizon training produces superlinear scaling $\hat{h} \propto h_{\mathrm{train}}^{2.07}$ below the $\hat{h}=h_{\mathrm{train}}$ diagonal ($H2$). At $h_{\mathrm{train}}=0$, all five seeds select $\hat{h}=0$. \emph{(b)} Post-switch good-to-bad detection ($k=12$ and $k=16$) and bad-to-good re-testing rates across $h_{\mathrm{train}}$, compared against the myopic ideal observer equipped with each agent's own $\hat{h}$ (open markers, dashed lines). Reactivity turns on sharply for $h_{\mathrm{train}}>0$ and plateaus at high volatility, with a localized shortfall at $k=16, h_{\mathrm{train}}=0.02$ ($0.54$ vs.\ $0.97$) reflecting soft policy thresholds near the episode horizon.}
\label{fig:sweep}
\end{figure}

\begin{table}[h]
\centering\small
\resizebox{\linewidth}{!}{%
\begin{tabular}{llll}
\toprule
& Prediction & Result (five seeds) & Verdict\\
\midrule
H1 & $\hat h$ increases with training $h$ (within seed) & Spearman $\rho=1$ in 5/5 seeds & holds\\
H2 & $\hat h$ proportional to $h$ (log--log slope interval includes 1) & slope 2.07 [1.95, 2.20] & \textbf{fails}\\
H3 & ideal observer with own $\hat h$ predicts detection (good$\to$bad) & mean abs.\ error 0.094 [0.072, 0.117]; bound 0.15 & holds\\
H4 & detection of good$\to$bad at $k=12$ increases with $h$ & $\rho>0$ in 4/5; mean 0.11 [0.04, 0.21] & holds\\
H5 & re-testing after bad$\to$good increases with $h$ & $\rho>0$ in 5/5; mean 0.98 [0.94, 1.00] & holds\\
\bottomrule
\end{tabular}}
\caption{\textbf{The second test.} $\hat h$ per agent in Table~\ref{tab:hz1seeds}. Intervals: seed bootstrap, 10,000 draws.}
\label{tab:hz1}
\end{table}

The rank ordering of $\hat{h}$ is exact in all five paired seeds (Fig.~\ref{fig:sweep}a and Table~\ref{tab:hz1}). Agents trained on stationary partners ($h_{\mathrm{train}}=0$) select $\hat{h}=0$, whereas at $h_{\mathrm{train}}=0.02$ the fitted values cluster at $0.001$--$0.003$, rising to $0.10$--$0.20$ at $h_{\mathrm{train}}=0.20$. While $\hat{h}$ increases monotonically with $h_{\mathrm{train}}$, it systematically underestimates slow switch rates, scaling superlinearly as $\hat{h} \propto h_{\mathrm{train}}^{2.07}$ (log--log slope $2.07$ [$1.95, 2.20$]) and thus falsifying linear proportionality ($H2$). This downward bias reflects finite-horizon episode statistics during reinforcement learning: at $h_{\mathrm{train}}=0.02$, the expected number of transitions per 20-round episode is only $19 \times 0.02 = 0.38$, so roughly $68\%$ of training episodes contain no switches at all, biasing the learned prior toward stationarity.

Figure~\ref{fig:sweep}b shows the resulting behavioral shift across the volatility sweep. When a good partner turns bad after round 12, detection jumps from $16\%$ at $h_{\mathrm{train}}=0$ to $91$--$96\%$ across all positive training hazards. For a very late switch after round 16, detection rises smoothly from $0\%$ at $h_{\mathrm{train}}=0$ through $54\%$ ($h_{\mathrm{train}}=0.02$) to a plateau of $85$--$88\%$ at higher hazards. Simultaneously, exploratory re-testing after a bad-to-good switch increases monotonically from $73\%$ to $96\%$, while earnings in stationary no-switch games decline only modestly from $22.14$ to $21.51$ per game.

Across all tested conditions, the myopic filter parameterized by each agent's own $\hat{h}$ predicts good-to-bad switch detection with a mean absolute error of $0.094$, satisfying the pre-registered $0.15$ bound ($H3$). However, Fig.~\ref{fig:sweep}b highlights a pronounced localized discrepancy near the end of the episode: at $h_{\mathrm{train}}=0.02$ and switch round $k=16$, the sharp-threshold Bayesian proxy predicts $97\%$ detection, whereas the trained agents achieve $54\%$. With only four rounds remaining after $k=16$, the network's finite-gain logistic readout maps noisy post-switch returns smoothly rather than crossing a discontinuous decision boundary.

To conclude this section, we have seen that training volatility sets how fast the agent forgets past evidence, though finite game length and noisy late draws prevent a single parameter from capturing every decision near the end of an episode.
\section{Intervening on reactivity}
\label{sec:intervention}

Varying the switch rate during training changes how readily the
learned policy responds to new evidence. Can we change that response
after training? The Bayesian filter gives us a quantity to measure:
how much a piece of evidence still affects the decision several
rounds later.

Fix the action sequence and write $e_s=a_s\ell(x_s)$ for the evidence
received at round $s$ and $u_s=\lambda_s+e_s$ for the posterior log-odds
before the transition step. Differentiating the filter gives the
response of the predictive belief at round $t$ to an earlier evidence
increment at round $\tau<t$:
\begin{equation}
K_h(t,\tau\mid\mathcal H)
\equiv \frac{\partial\lambda_t}{\partial e_\tau}
= \prod_{s=\tau}^{t-1} g_h'(u_s),
\qquad
g_h'(u)=
\frac{1-2h}{1+2h(1-h)(\cosh u-1)} .
\label{eq:belief-response}
\end{equation}
Here $\mathcal H$ is the transcript along which the response is
evaluated. For $h=0$, every factor is one: an evidence increment
retains its full weight, regardless of when it arrived. For
$0<h<1/2$, when the intermediate log-odds $u_s$ stay close to zero,
the response depends approximately only on the lag
$\Delta t=t-\tau$:
\begin{equation}
K_h(\Delta t)\simeq (1-2h)^{\Delta t}
= \exp[-\Delta t/\tau_{\mathrm{mem}}],
\qquad
\tau_{\mathrm{mem}}=-\frac{1}{\log(1-2h)}
\simeq \frac{1}{2h}
\quad (h\ll 1).
\label{eq:memory-time}
\end{equation}
Along a general history, large $|u_s|$ reduce $g_h'(u_s)$ further.
The effect of an old return therefore depends on both its age and
the evidence encountered since then.

We tested sensitivity to evidence order using paired transcripts.
We kept the action sequence fixed and exchanged two observed returns,
one early and one late. The two transcripts contain the same total
log-likelihood evidence, so the stationary filter assigns them
identical beliefs at the final decision. A filter with $h>0$ can
assign them different beliefs. For each pair we measured the change
$\Delta z$ in investment log-odds when the return that provides
stronger evidence for a good partner is placed at the recent
position rather than the earlier one. We computed the predicted
$\Delta z$ using the full recursion and the previously fitted
choice model for each policy.

\begin{figure}[t]
\centering
\includegraphics[width=\linewidth]{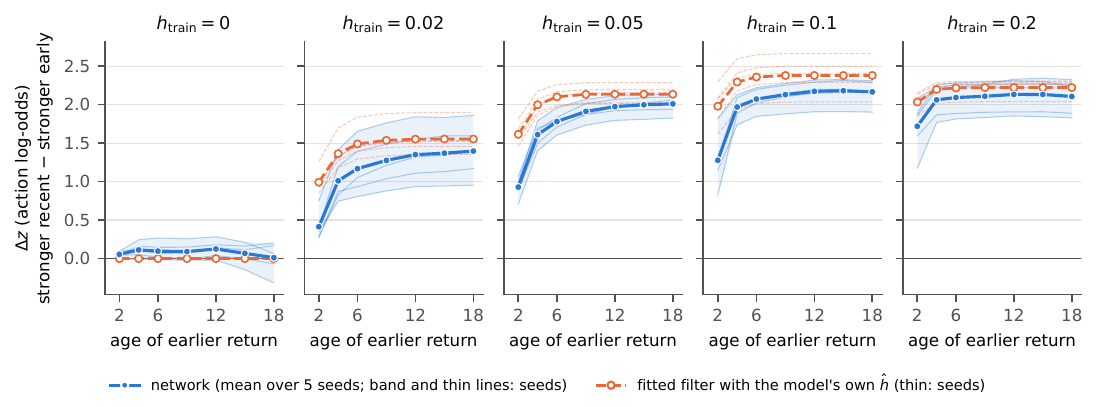}
\caption{\textbf{Sensitivity to the order of evidence.}
Each panel shows one training hazard $h_{\rm train}$. The two
transcripts in each pair have identical actions and returns, with
two returns exchanged. The recent position is fixed and the age
of the earlier return is varied. Positive $\Delta z$ means that
investment is more likely when the stronger evidence for a good
partner is placed at the recent position. Blue shows the network
response and orange shows the prediction of the fitted filter
using each policy's own $\hat h$. Thick curves are means over
five seeds. Thin curves show individual seeds, and the blue band
shows their range.}
\label{fig:swap}
\end{figure}

The stationary-trained policies are nearly invariant to the swap
(Fig.~\ref{fig:swap}). In every positive training-hazard condition,
the policies are more likely to invest when the stronger evidence
is recent. The effect grows as the earlier return is moved further
back and then levels off. The fitted Bayesian filter captures this
broad pattern, but overestimates the effect of swapping nearby
returns. The agreement is closer at larger separations.

The same pairs provide a test of a post-training intervention.
Measure the policy's response to exchanging returns before and
after the intervention, and compare the shift with the predictions
for different hazards. An intervention that increases reactivity
should make older evidence count less relative to recent evidence.
A constant shift in investment log-odds cancels in the paired
difference.

In the activation experiment reported here, perturbing the
layer-14 probe direction had little immediate effect on choice.
After removal of that projection, a refitted probe still recovered
97\% of the original improvement in log loss over chance at
layer~14. The swap experiment gives us a baseline for testing
whether a future intervention changes how the policy weights
evidence over time.

\section{Discussion}

A small language model trained by reward learns both effective play and an internal state representation that tracks hidden partner transitions near the Bayesian bound. Whether that internal representation drives rapid behavioral adaptation, however, is governed by the volatility of the training environment. The constant-hazard Bayesian filter captures this separation through a single dynamical parameter: at $h=0$, extensive evidence accumulation produces deep hysteresis and lock-in, whereas $h>0$ imposes a saturation ceiling $\lambda_{\max}(h) = \ln[(1-h)/h]$ that bounds the memory horizon and amplifies recent contradictory evidence.

The inferred hazard rate $\hat{h}$ provides a single effective parameter that tracks how training volatility shapes policy reactivity. At the same time, the specific ways in which the network departs from the minimal Bayesian filter reveal the inductive biases of a finite-capacity sequence model: imperfect memory retention even when trained without switches ($h_{\mathrm{train}}=0$), under-reaction to rare switches in finite 20-round episodes ($\hat{h} \propto h_{\mathrm{train}}^{2.07}$), and softer decision boundaries near the end of the game. 

Finally, because partner identity is encoded redundantly across many activation directions, perturbing a single linear probe direction leaves behavior unchanged. Future causal interventions on reactivity should therefore target how the network weights past evidence over time, hence directly measuring shifts in its temporal memory kernel $K(\Delta t)$ rather than static state readouts.

\paragraph{Acknowledgements.} The author would like to acknowledge conversations with AI systems from Anthropic and OpenAI while this study was developed. The author used agents in Claude to write and run the code.  The author acknowledges the computer resources at Artemisa and the technical support provided by the Instituto de Fisica Corpuscular, IFIC(CSIC-UV). Artemisa is co-funded by the European Union through the 2014-2020 ERDF Operative Programme of Comunitat Valenciana, project IDIFEDER/2018/048.

\bibliographystyle{plainnat}
\bibliography{refs}
\clearpage
\begin{appendices}
\section{The game and Bayesian reference}
\label{app:theory}

The training means are $1/3\pm0.1$ and $1/3\pm0.2$. The test means are $1/3\pm0.15$. For both, $X\sim\mathrm{Beta}(c\kappa,(1-c)\kappa)$ with $\kappa=2\times 25^{0.9}\simeq36.2$. The observed payoff $3X$ is rounded to two decimals. For an unrounded draw the log-likelihood ratio is
\[
 \ell(x)=\log\!\frac{B(c_B\kappa,(1-c_B)\kappa)}{B(c_G\kappa,(1-c_G)\kappa)}
 +\kappa(c_G-c_B)\log\frac{x}{1-x}.
\]
For a displayed payoff the exact likelihood integrates over its rounding bin. The two expected log-likelihood increments have opposite signs, with $\mathbb E_G[\ell]=7.35\text{ nats}$ and $\mathbb E_B[\ell]=-8.80\text{ nats}$ for the test partners. With always-investing play the drift approximation in Eq.~\eqref{eq:e2} places the transition around switch round 9 for detection near the end of the game. Finite return fluctuations spread this crossover over several rounds.

The two groups in the original switch experiment used stationary training or a single switch in half of training games. For that comparison the Bayesian change-point reference assumes prior probability $1/2$ of one switch, uniformly placed after rounds 1 through 19. The reference averages over all compatible switch times. The constant-hazard recursion in Eq.~\eqref{eq:e1} allows repeated switches and is the reference for the hazard sweep. Both references start with equal prior probabilities of good and bad. A myopic player invests when the predicted return exceeds one coin ($\lambda_t > 0$). A player that optimises future as well as immediate return can also invest to acquire information.

\section{Training, evaluation and statistical summaries}
\label{app:training}

Each policy starts from pretrained Qwen3. The decision is the softmax restricted to the A and B tokens at the end of the transcript. The adapter rank is 64 and its scale parameter is 16. We train for 1,000 updates with 16 games per update and 20 rounds per game. REINFORCE uses undiscounted return-to-go, an exponential moving baseline for each round, batch-normalised advantages and an entropy bonus of 0.1. The optimiser is AdamW with learning rate $10^{-4}$, 5\% warm-up and cosine decay. The scrambled control deranges rewards across games within each batch and round for the learning update.

In the original switching evaluation, each policy plays 128 games per schedule and seed. Schedules include stationary good and bad partners and switches in either direction after rounds 5, 10 and 15. For each switch schedule, the adaptation index is
\[
 A=\frac{\bar x_{\rm switch}-\bar x_{\rm old}}{\bar x_{\rm new}-\bar x_{\rm old}},
\]
with the bars denoting means over the same post-switch window. The index can lie outside $[0,1]$ and becomes unstable when the baseline difference is small. The final experiment uses switches after rounds 4, 8, 12 and 16. Detection denotes a mean investment probability in rounds 18--20 on the side of $1/2$ appropriate to the new partner. Re-testing denotes at least one investment in the first five rounds after a bad-to-good switch.

Confidence intervals use 10,000 percentile-bootstrap draws. Comparisons across models resample paired training seeds. Where both histories and seeds enter the uncertainty calculation, the bootstrap resamples these two units independently and preserves common evaluation draws across conditions. Five seeds give limited resolution for uncertainty across training runs. We report effect sizes and per-seed data for the main tests below.

\section{Read-outs and activation interventions}
\label{app:steer}

The stationary-type probe is a regularised logistic regression on standardised residual-stream activations at the decision position. It fits 400 random-action training histories and scores 400 held-out histories. Each model sees the same transcripts. The current-type probe uses switching histories and scores the instantaneous partner type before each decision on novel histories. Layer 24 was selected by a fixed rule on independent models.

We steer along the unit vector $u$ of the stationary-type probe in raw activation coordinates at layer 14, scaled by its projected training standard deviation. The primary comparison contrasts the change in action log odds under opposite two-standard-deviation shifts with that for 20 random directions. The mean excess response is $5.1\times10^{-4}$ log-odds units. The matched ablation replaces the projection along $u$ by its training mean at every position. A supplementary check fits fresh probes to the resulting activations. The retained fraction of the log-loss improvement over chance is 0.97 at layer 14 and 1.00 to reported precision at layers 24 and 28.

\begin{table}[h]
\centering\small
\resizebox{\linewidth}{!}{%
\begin{tabular}{lccc}
\toprule
& Layer 14 & Layer 24 & Layer 28\\
\midrule
$L_{\mathrm{none}}$, per seed & 0.053, 0.064, 0.075, 0.064, 0.083 & 0.051--0.053 & 0.052--0.055\\
$L_u$ (refitted decoder), per seed & 0.055, 0.079, 0.096, 0.087, 0.121 & 0.051--0.053 & 0.052--0.055\\
$L$ with a random direction removed, change & $\le0.0002$ & $\le0.0003$ & $\le0.0008$\\
\midrule
$R$, refitted decoder & \textbf{0.97} [0.94, 0.99] & \textbf{1.00} [1.00, 1.00] & \textbf{1.00} [1.00, 1.00]\\
Reading & remains decodable & remains decodable & remains decodable\\
$R$, original decoder & $-0.05$ [$-0.08$, $-0.03$] & 1.00 [1.00, 1.00] & 1.00 [1.00, 1.00]\\
\bottomrule
\end{tabular}}
\caption{Decoder test log loss (nats per decision) after removing the decoded direction $u$ at layer 14. The original layer-14 decoder fails after the removal, as it must, since it reads exactly the removed component. A newly fitted decoder recovers most of the improvement in log loss over chance, and the later layers are unaffected. ``Change'' is the largest absolute difference from $L_{\mathrm{none}}$ over the 20 random directions and five seeds. Untrained model, for reference: $L_{\mathrm{none}}=0.124$ and $L_u=0.165$ at layer 14. Exact Bayesian floor: 0.048.}
\label{tab:ad1}
\end{table}

The auxiliary decision head receives one frozen residual-stream vector at the decision position. A linear encoder parameterises the mean and log variance of an eight-dimensional Gaussian latent. The action probability is a probit applied to its linear read-out and averaged analytically over that latent. We train the head with REINFORCE and a KL penalty of 0.01 relative to the unit Gaussian, using the same game budget as for the original policies. The layer-28 result illustrates the instability of the adaptation index: two heads reverse the stationary good--bad baseline. The descriptive absolute contrast remains defined for these heads.

\begin{table}[h]
\centering\small
\setlength{\tabcolsep}{3pt}
\resizebox{\linewidth}{!}{%
\begin{tabular}{lccccc}
\toprule
& \multicolumn{2}{c}{Invest rate} & & Reward per round & Reward per game\\
\cmidrule(lr){2-3}
& no switch & after switch & Follows & after switch & with / without\\
& good / bad & g$\to$b / b$\to$g & the switch & g$\to$b / b$\to$g & a switch\\
\midrule
Stationary-trained agent (own choices) & 0.81 / 0.21 & 0.60 / 0.33 & $-0.26$ & 0.73 / 1.15 & 20.5 / 22.3\\
Head, layer 14 & 0.65 / 0.42 & 0.46 / 0.56 & $+0.09$ & 0.79 / 1.25 & 20.6 / 20.9\\
Head, layer 24 & 0.67 / 0.41 & 0.49 / 0.57 & $+0.08$ & 0.78 / 1.26 & 20.7 / 21.0\\
Switch-trained agent & 0.79 / 0.30 & 0.39 / 0.64 & $+0.25$ & 0.82 / 1.29 & 21.6 / 22.0\\
\midrule
Ignores the partner & 0.58 / 0.58 & 0.58 / 0.58 & \phantom{+}0.00 & 0.74 / 1.26 & 20.0 / 20.0\\
Oracle & 1 / 0 & 0 / 1 & $+1$ & 1.00 / 1.45 & 24.5 / 24.5\\
\bottomrule
\end{tabular}}
\caption{\textbf{Post-switch behaviour and rewards} (means over the three switch rounds and five seeds). ``Follows the switch'' is the invest rate after a switch to a good partner minus that after a switch to a bad one: positive values favour the current partner, negative values the old one; its per-seed values are $-0.24$ to $-0.30$ (stationary-trained agent), $+0.04$ to $+0.15$ (layer-14 heads), $-0.03$ to $+0.15$ (layer-24 heads) and $+0.21$ to $+0.27$ (switch-trained agents). ``Ignores the partner'' is the untrained model; the oracle knows the current type. Rewards are observed payoffs per round in rounds after the switch, and per 20-round game; oracle values are expected payoffs. Descriptive.}
\label{tab:postswitch}
\end{table}

\section{Predicting choices from Bayesian beliefs}
\label{app:bt1}

The first test uses the stationary and one-switch models on fresh transcripts with random actions, plus on-policy games with switches after rounds 7 and 12. For each of 32 transcript schedules, we fit on 25 histories and evaluate on a separate 25. The soft target is the model's investment probability. Each logistic fit takes the belief log odds clipped to $\pm10$ as its input. The score is cross-entropy above the entropy of the model's own probability. For the belief read-out we fit ridge regression from standardised activations to the ideal belief probability and report $R^2$ on held-out histories. The regularisation is selected by cross-validation on the fitting histories.

\begin{table}[h]
\centering\small
\resizebox{\linewidth}{!}{%
\begin{tabular}{lccccc}
\toprule
Seed & 3100 & 3101 & 3102 & 3103 & 3104\\
\midrule
\multicolumn{6}{l}{\emph{Stationary-trained}}\\
$\Delta$ (T1a) & +0.087 & +0.098 & +0.093 & +0.094 & +0.107\\
$R^2$ layer 24, $h=0$ / change-point & 0.87 / 0.95 & 0.86 / 0.95 & 0.87 / 0.93 & 0.89 / 0.95 & 0.86 / 0.92\\
Detect good$\to$bad, $k=7$ / 12 & 0.98 / 0.70 & 0.99 / 0.34 & 0.54 / 0.01 & 0.94 / 0.12 & 1.00 / 0.63\\
Detect bad$\to$good, $k=7$ / 12 & 0.19 / 0.02 & 0.46 / 0.04 & 0.71 / 0.09 & 0.63 / 0.13 & 0.11 / 0.01\\
\multicolumn{6}{l}{\emph{Switch-trained}}\\
$\Delta$ (T1b) & $-0.026$ & $-0.051$ & $-0.047$ & $-0.045$ & $-0.054$\\
$R^2$ layer 24, $h=0$ / change-point & 0.83 / 0.97 & 0.82 / 0.98 & 0.81 / 0.93 & 0.84 / 0.96 & 0.82 / 0.97\\
Detect good$\to$bad, $k=7$ / 12 & 0.98 / 0.99 & 1.00 / 0.99 & 0.98 / 0.99 & 0.98 / 0.98 & 0.99 / 1.00\\
Detect bad$\to$good, $k=7$ / 12 & 1.00 / 0.96 & 1.00 / 0.92 & 0.98 / 0.94 & 0.99 / 0.88 & 0.98 / 0.89\\
\bottomrule
\end{tabular}}
\caption{BT1 per seed. Untrained model (descriptive): $\Delta\approx0$; $R^2$ at layer 24, 0.63 ($h=0$) and 0.77 (change-point). Compute: 2.2 GPU-hours plus 0.2 for the engineering check.}
\end{table}

\section{The training hazard sweep}
\label{app:hz1}

The sweep changes the training process at each round boundary: with probability $h_{\rm train}$ the partner is replaced with one in the other category. We fit the constant-hazard Bayesian filter on the random-action histories separately for each of the 25 agents. Candidate hazards are $0$, $0.001$, $0.003$, $0.01$, $0.02$, $0.03$, $0.05$, $0.075$, $0.10$, $0.15$, $0.20$, $0.30$ and $0.40$. For each candidate we use the logistic fit described above. The minimiser of the fitting loss gives $\hat h$. A myopic player with that hazard then plays separate evaluation games using the same underlying draws as the learned policy.

\begin{table}[h]
\centering\small
\resizebox{\linewidth}{!}{%
\begin{tabular}{lccccc}
\toprule
Training $h$ & 0 & 0.02 & 0.05 & 0.1 & 0.2\\
\midrule
\multicolumn{6}{l}{\emph{Effective hazard $\hat h$, seeds 3400--3404}}\\
3400 & 0 & 0.001 & 0.02 & 0.075 & 0.2\\
3401 & 0 & 0.001 & 0.01 & 0.03 & 0.15\\
3402 & 0 & 0.001 & 0.03 & 0.05 & 0.15\\
3403 & 0 & 0.001 & 0.02 & 0.03 & 0.1\\
3404 & 0 & 0.003 & 0.01 & 0.075 & 0.2\\
\multicolumn{6}{l}{\emph{Means over seeds: agents (ideal observer with own $\hat h$)}}\\
Detect good$\to$bad, $k=4$ & 0.92 (1.00) & 0.99 (1.00) & 0.99 (1.00) & 0.95 (1.00) & 0.89 (1.00)\\
Detect good$\to$bad, $k=8$ & 0.71 (0.95) & 0.99 (1.00) & 0.98 (1.00) & 0.95 (1.00) & 0.92 (1.00)\\
Detect good$\to$bad, $k=12$ & 0.16 (0.06) & 0.96 (1.00) & 0.95 (1.00) & 0.94 (1.00) & 0.91 (1.00)\\
Detect good$\to$bad, $k=16$ & 0.00 (0.02) & 0.54 (0.97) & 0.85 (0.99) & 0.88 (0.99) & 0.86 (0.99)\\
Detect bad$\to$good, $k=8$ / 16 & 0.39 / 0.01 & 0.97 / 0.32 & 0.99 / 0.53 & 1.00 / 0.69 & 1.00 / 0.94\\
Re-test (pooled $k$) & 0.73 & 0.79 & 0.87 & 0.91 & 0.96\\
Reward, no-switch games & 22.14 & 22.12 & 21.95 & 21.90 & 21.51\\
$R^2$ layer 24, $b_0$ / $b_{h}$ & 0.87 / 0.87 & 0.83 / 0.97 & 0.82 / 0.98 & 0.81 / 0.98 & 0.81 / 0.98\\
\bottomrule
\end{tabular}}
\caption{HZ1 per agent and per level. Compute: 81.4 GPU-hours of training and 5.2 of evaluation, plus 0.3 for the engineering check.}
\label{tab:hz1seeds}
\end{table}

\section{Order of analyses}

The stationary decoding, switch comparison, probe intervention and decision-head studies were completed before the Bayesian account was written. The post-ablation decoding check and the absolute decomposition of the head results were added during revision. The two theory tests specify fresh data and decision rules. In the first test all five predictions were required by the original rule and the lock-in prediction failed. In the sweep, ordering of fitted hazards, accuracy of the good-to-bad detection benchmark and the round-12 detection contrast were the required tests. They passed. The prediction of proportional fitted and training hazards failed. Protocols and freeze records were committed to the project repository before the corresponding tests. These records document the order of analyses within the project.
\end{appendices}
\end{document}